\documentclass[a4paper,twocolumn,american,english,conference,a4paper,10p]{IEEEtran}
\usepackage[T1]{fontenc}
\usepackage[latin9]{inputenc}
\usepackage{array}
\usepackage{units}
\usepackage{textcomp}
\usepackage{bbding}
\usepackage{multirow}
\usepackage{amsbsy}
\usepackage{amstext}
\usepackage{amssymb}
\usepackage{graphicx}

\makeatletter

\pdfpageheight\paperheight
\pdfpagewidth\paperwidth

\providecommand{\tabularnewline}{\\}

\usepackage{cite}
\usepackage{amsmath,amssymb,amsfonts}
\usepackage{algorithmic}
\usepackage{graphicx}
\usepackage{textcomp}
\usepackage{xcolor}

\usepackage{booktabs}

\usepackage{textcomp}
\usepackage{stfloats}
\usepackage{url}
\usepackage{verbatim}
\usepackage{graphicx}
\usepackage{balance}
\usepackage{adjustbox}
\usepackage{mathtools}
\usepackage{xspace}

\usepackage{titlesec}
\titleformat{\paragraph}[runin]{\itshape\normalsize\mdseries}{\theparagraph}{1em}{}
\titlespacing*{\paragraph}{0pt}{1ex plus 1ex minus .2ex}{1em}

\usepackage{environ}         
\usepackage{etoolbox}        

\usepackage[pagebackref,breaklinks,colorlinks]{hyperref}

\newlength{\myl}
\let\origequation=\equation
\let\origendequation=\endequation

\RenewEnviron{equation}{
  \settowidth{\myl}{$\BODY$}                       
  \origequation
  \ifdimcomp{0.9\linewidth}{>}{\the\myl}
  {\ensuremath{\displaystyle\BODY}}                             
  {\resizebox{0.9\linewidth}{!}{\ensuremath{\displaystyle\BODY}}}  
  \origendequation
}

\makeatother

\usepackage{babel}
\begin{document}
\selectlanguage{american}%



\global\long\def\imSymbol{I}%
 
\global\long\def\imWidth{W}%
 
\global\long\def\imHeight{H}%
 
\global\long\def\hImCoord{i}%
 
\global\long\def\vImCoord{j}%


\global\long\def\focal#1{f_{#1}}%


\global\long\def\percentile#1{\text{P}_{#1}}%

\global\long\def\pointP{\text{P}}%

\global\long\def\transformationMatrix#1{\text{\textbf{T}}_{#1}}%

\global\long\def\rotMatrix#1{\mathbf{R}_{#1}}%

\global\long\def\transpositionSymbol{T}%

\global\long\def\translationVector{\mathbf{\mathbf{t}}}%

\global\long\def\translationVectorComp{\mathrm{t}}%

\global\long\def\time{t}%

\global\long\def\indexOfVirtualCamera{V}%


\global\long\def\dmap{\mathbf{d}}%
 
\global\long\def\dpix{\text{z}}%

\global\long\def\disp{\rho}%
 
\global\long\def\dispmap{\boldsymbol{\rho}}%
\global\long\def\dispmapEstimate{\boldsymbol{{\hat{\rho}}}}%

\global\long\def\umap{\mathbf{u}}%
 
\global\long\def\upix{\text{u}}%

\global\long\def\omap{\mathbf{o}}%
 
\global\long\def\opix{\text{o}}%

\global\long\def\firstIndex{\time-1}%

\global\long\def\secondIndex{\time}%

\global\long\def\location{\mu}%

\global\long\def\scale{b}%

\global\long\def\dda{a}%

\global\long\def\ddb{c}%

\global\long\def\facdepth{\delta_{\dpix}}%

\global\long\def\facdisp{\delta_{\disp}}%


\global\long\def\upscaled#1{\boldsymbol{\hat{\ }}#1}%

\global\long\def\levelIndex{l}%


\global\long\def\comma{\,,}%

\global\long\def\pointMath{\,.}%

\global\long\def\unreal{\textrm{the Unreal engine}}%


\global\long\def\comma{\,\mbox{,}}%

\global\long\def\dot{\,\mbox{.}}%

\newcommand{\eqx}[1]{equation\ (#1)\xspace}               
\newcommand{\eqxy}[2]{equations\ (#1) and\ (#2)\xspace}   
\newcommand{\eqxyz}[3]{equations\ (#1),\ (#2), and\ (#3)\xspace}   
\newcommand{\Eqx}[1]{Eq.\ (#1)\xspace}                   
\newcommand{\Eqxy}[2]{Eq.\ (#1) and\ (#2)\xspace}        
\newcommand{\Eqxyz}[3]{Eq.\ (#1),\ (#2), and\ (#3)\xspace}        

\newcommand{\figx}[1]{Figure\ #1\xspace}                  
\newcommand{\figxy}[2]{Figures\ #1 and\ #2\xspace}        
\newcommand{\Figx}[1]{Fig.\ #1\xspace}                    
\newcommand{\Figxy}[2]{Fig.\ #1 and\ #2\xspace}           
\newcommand{\tabx}[1]{Table\ #1\xspace}                   
\newcommand{\tabxy}[2]{Tables\ #1 and #2\xspace}          

\newcommand{\secx}[1]{Section\ #1\xspace}                 
\newcommand{\secxy}[2]{Sections\ #1 and\ #2\xspace}       
\newcommand{\secxyzt}[4]{Sections\ #1,\ #2,\ #3, and\ #4\xspace}
\newcommand{\chapx}[1]{Chapter\ #1\xspace}                
\newcommand{\chapxy}[2]{Chapters\ #1 and #2\xspace}                

\DeclareRobustCommand\onedot{\futurelet\@let@token\@onedot}
\def\@onedot{\ifx\@let@token.\else.\null\fi\xspace}

\def\eg{\emph{e.g.}\xspace}
\def\Eg{\emph{E.g.}\xspace}
\def\ie{\emph{i.e.}\xspace}
\def\Ie{\emph{I.e.}\xspace}
\def\cf{\emph{c.f.}\xspace}
\def\Cf{\emph{C.f.}\xspace}
\def\etc{\emph{etc.}\xspace}
\def\vs{\emph{vs.}\xspace}
\def\wrt{w.r.t.}
\def\dof{d.o.f.}
\def\etal{\emph{et al.}\xspace}

\newcommand{\kitti}{KITTI\xspace}
\newcommand{\tartanair}{TartanAir\xspace}
\newcommand{\midair}{Mid-Air\xspace}
\newcommand{\midairgithub}{\url{https://github.com/michael-fonder/M4Depth}}
\newcommand{\mdupara}{M4Depth+U$ _\disp$\xspace}
\newcommand{\mdudepth}{M4Depth+U$ _\dpix$\xspace}\selectlanguage{english}%

\title{A technique to jointly estimate depth and depth uncertainty for unmanned aerial vehicles
\thanks{This work was partly supported by the Walloon region (Service Public de Wallonie Recherche, Belgium) under grant n°2010235 (ARIAC by DigitalWallonia.ai)}
}
\author{
    \IEEEauthorblockN{         Michaël Fonder$^1$, and Marc Van Droogenbroeck$^1$     }
    \IEEEauthorblockA{
        \\
        $^1$ Montefiore Institute, University of Liège, Liège, Belgium \\
		\\
        \texttt{michael.fonder@uliege.be}
    }
}
\maketitle
\begin{abstract}
When used by autonomous vehicles for trajectory planning or obstacle
avoidance, depth estimation methods need to be reliable. Therefore,
estimating the quality of the depth outputs is critical. In this paper,
we show how M4Depth, a state-of-the-art depth estimation method designed
for unmanned aerial vehicle (UAV) applications, can be enhanced to
perform joint depth and uncertainty estimation. For that, we present
a solution to convert the uncertainty estimates related to parallax
generated by M4Depth into uncertainty estimates related to depth,
and show that it outperforms the standard probabilistic approach.
Our experiments on various public datasets demonstrate that our method
performs consistently, even in zero-shot transfer. Besides, our method
offers a compelling value when compared to existing multi-view depth
estimation methods as it performs similarly on a multi-view depth
estimation benchmark despite being 2.5 times faster and causal, as
opposed to other methods. The code of our method is publicly available
at the following URL: \url{https://github.com/michael-fonder/M4DepthU}.
\end{abstract}


\section{Introduction\label{sec:Introduction}}

One of the many applications of depth estimation is to replace depth
sensors in autonomous vehicles for path planning~\cite{Mumuni2022Deep}
or obstacle avoidance~\cite{Yang2019Reactive,Wang2020UAVEnvironmental}.
Such practice is common for small unmanned aerial vehicles (UAVs)
as their size, weight and power constraints prevent the use of dedicated
depth sensors. For such applications, being able to predict the quality
of the estimates is essential to anticipate potentially erroneous
data and take action accordingly. However, to the best of our knowledge,
the task of joint depth and uncertainty estimation for drone-specific
constraints, such as being robust to a wide variety of conditions
and environments while being computationally lightweight enough to
run in real-time on limited hardware, has not been addressed yet.

In a previous work~\cite{Fonder2022Parallax}, we introduced M4Depth,
a depth estimation method specifically designed for unstructured environments
and UAV applications that shows state-of-the-art performance for depth
estimation in such environments and in generalization. In this work,
we detail how it is possible to adapt the architecture of M4Depth
to jointly estimate depth and its uncertainty for a negligible additional
computational cost. Section~\ref{sec:Related-work} discusses the
related works about uncertainty estimation. In Section~\ref{sec:Uncertainty},
we explain how M4Depth can be adapted for joint depth and uncertainty
estimation. Our experiments, presented in Section~\ref{sec:Experiments},
test our proposal in various conditions including zero-shot transfer
on public datasets and on an existing benchmark for multi-view depth
(MVD) estimation methods. Section~\ref{sec:Conclusion} concludes
this work.

\begin{figure}
\begin{adjustbox}{width=0.99\linewidth}%
\noindent\begin{minipage}[t]{1\linewidth}%
\setlength{\tabcolsep}{2pt}%
\begin{tabular}{ccc}
\includegraphics[width=0.32\textwidth]{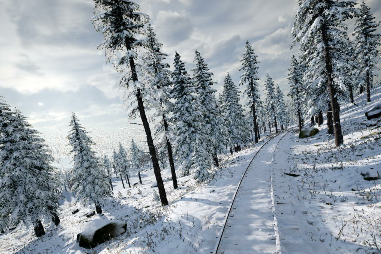} & \includegraphics[width=0.32\textwidth]{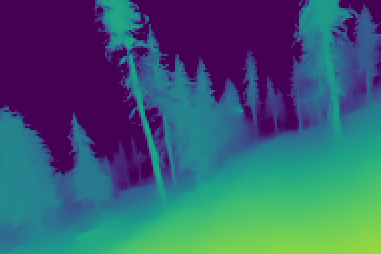} & \includegraphics[width=0.32\textwidth]{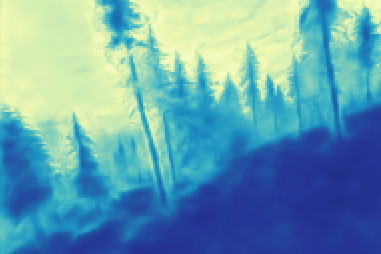}\tabularnewline
\includegraphics[width=0.32\textwidth]{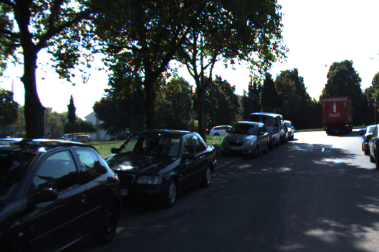} & \includegraphics[width=0.32\textwidth]{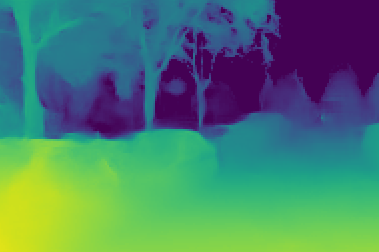} & \includegraphics[width=0.32\textwidth]{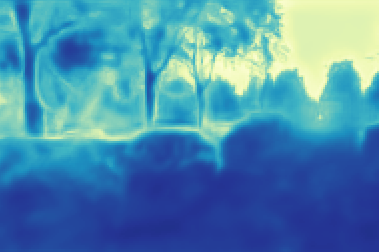}\tabularnewline
\vspace{-0.4cm}
 &  & \tabularnewline
{\footnotesize{}RGB input} & {\footnotesize{}Estimated depth} & {\footnotesize{}Estimated uncertainty}\tabularnewline
\end{tabular}%
\end{minipage}\end{adjustbox}\caption{Illustration of depth and uncertainty estimates produced by the method
presented in this work for two setups. Row~1: trained and tested
on the \midair~\cite{Fonder2019MidAir} dataset. Row~2: tested
in zero-shot transfer on the \kitti~\cite{Geiger2012AreWe} dataset.
Lighter colors correspond to higher uncertainty values. \label{fig:graphical_abstract}}
\end{figure}

Our main contributions are as follows. (i) Our method is the first
to address joint monocular depth and uncertainty estimation for the
specific constraints of autonomous drones. (ii) We show that our method
for uncertainty estimation performs consistently in zero-shot transfer
in different environments. (iii) On a benchmark for MVD, we show that
our method performs on par with existing MVD methods for joint depth
and uncertainty estimation despite being $2{.}5$ times faster and
causal, as opposed to MVD methods.

\section{Related works\label{sec:Related-work}}

Our M4Depth paper~\cite{Fonder2022Parallax} already covers related
works in depth estimation, and uncertainty in neural networks is well
covered in the survey of Gawlikowski~\etal~\cite{Gawlikowski2021ASurvey}.
Therefore, we focus on uncertainty estimation for pixel-wise computer
vision regression tasks in this section.

Kendall and Gal~\cite{Kendall2017What} showed that a part of the
uncertainty in a deep neural network, called the \emph{aleatoric}
uncertainty, is due to the noise in the input data. They also showed
that this part of the uncertainty can be estimated by training a network
to learn the parameters of a probabilistic distribution that represents
the noise in the output. The output noise for a method trained with
a L1 loss is assumed to follow a Laplace distribution~\cite{Ke2021Deep,Ilg2018Uncertainty,Schroppel2022ABenchmark,Zhang2022VisMVSNet}
as it allows learning its parameters, the location and the scale,
with a Maximum Log-Likelihood loss function~\cite{Kendall2017What,Ilg2018Uncertainty,Poggi2020On}.

Estimating the aleatoric uncertainty can be done either by creating
a new architecture designed around uncertainty, such as done by Ke~\etal~\cite{Ke2021Deep}
and Su~\etal~\cite{Su2022Uncertainty}, or by modifying an existing
architecture for the desired task. In the latter case, the simplest
way to proceed consists of adding a channel for the uncertainty at
the output of the network~\cite{Poggi2020On,Liu2019Neural,Zhao2021AConfidenceBased,Mehltretter2021Aleatoric,Zhang2022VisMVSNet}.
However, some methods create a distinct head for the uncertainty by
duplicating the last layers of their network~\cite{Homeyer2022Multiview,Klodt2018Supervising,Yang2021Fast},
which provides more trainable parameters to avoid potentially sub-optimal
performances due to the shared weights.

\section{Uncertainty estimation using M4Depth\label{sec:Uncertainty}}

In this section, we briefly remind the working principles of M4Depth~\cite{Fonder2022Parallax}
and explain how the network architecture can be modified to jointly
estimate the parallax and its aleatoric uncertainty. We then detail
how to get the uncertainty on depth from the uncertainty on the parallax.

\subsection{M4Depth working principles}

\noindent M4Depth is a multi-level pyramidal architecture where each
level has the same structure and outputs a parallax estimate. The
parallax $\disp>0$ is linked to the depth $\dpix$ of a point $\pointP$
by the motion of the camera between two poses:
\begin{equation}
\dpix=\frac{\sqrt{\left(\focal x\translationVectorComp_{x}-\translationVectorComp_{z}\hImCoord_{\indexOfVirtualCamera}\right)^{2}+\left(\focal y\translationVectorComp_{y}-\translationVectorComp_{z}\vImCoord_{\indexOfVirtualCamera}\right)^{2}}}{\disp\ \dpix_{\indexOfVirtualCamera}}-\frac{\translationVectorComp_{z}}{\dpix_{\indexOfVirtualCamera}}\comma\label{equ:disp_to_depth-reminder}
\end{equation}
where $\focal x$ and $\focal y$ are the respective focal lengths
along the $x$ and $y$ camera axes, $\left[\begin{array}{ccc}
\translationVectorComp_{x} & \translationVectorComp_{y} & \translationVectorComp_{z}\end{array}\right]$ expresses the known translation of the camera between the two poses,
and where $\hImCoord_{\indexOfVirtualCamera}$, $\vImCoord_{\indexOfVirtualCamera}$
and $\dpix_{\indexOfVirtualCamera}$ are solely functions of the projection
coordinates $(\hImCoord,\vImCoord)$ of $\pointP$ and the rotation
of the camera between the two poses~\cite{Fonder2022Parallax}.

\noindent The network starts with a first rough low-resolution parallax
estimate and then refines it progressively at higher resolutions to
get the final estimate. Each intermediate parallax map can be converted
into a depth map using \Eqx{\ref{equ:disp_to_depth-reminder}} for
each pixel. The only architectural modification required for joint
uncertainty inference is to add an output for uncertainty at each
level of the architecture. As for the parallax, the additional outputs
are refined progressively to get the final estimate. Details on the
architecture modifications can be found in our code.

\noindent As mentioned in~\cite{Fonder2022Parallax}, M4Depth is
trained for depth estimation on a weighted sum of the $L_{1}$ distance
of the logarithm of the depth for each architecture level $\levelIndex$:
\begin{equation}
\mathcal{L}_{t}=\frac{1}{\imHeight\imWidth}\sum_{\levelIndex=1}^{M}\sum_{\dpix_{ij}\in\dmap_{t}^{l}}2^{-\levelIndex}\ \left|\log(\dpix_{ij})-\log(\hat{\dpix}_{ij})\right|\pointMath\label{eq:L1_loss_depth}
\end{equation}

\subsection{Correspondence between depth and parallax uncertainties}

Since M4Depth works with parallax instead of depth values, we need
to make some adaptations to get the uncertainty on depth. We first
detail the baseline approach, which relies on the standard probabilistic
framework, to get depth uncertainties from M4Depth. We then present
a new and more elaborate method to get depth uncertainty estimates
from the parallax ones. As confirmed by experiments, our new method
better evaluates the depth uncertainty.

To simplify the notations in the following, we rewrite \Eqx{\ref{equ:disp_to_depth-reminder}}
for a given pixel and a given camera motion as:
\begin{equation}
\dpix=Z(\disp)=\frac{\dda}{\disp}+\ddb\comma\label{equ:disp_to_depth-simplified}
\end{equation}
where 
\begin{equation}
\dda=\frac{\sqrt{\left(\focal x\translationVectorComp_{x}-\translationVectorComp_{z}\hImCoord_{\indexOfVirtualCamera}\right)^{2}+\left(\focal y\translationVectorComp_{y}-\translationVectorComp_{z}\vImCoord_{\indexOfVirtualCamera}\right)^{2}}}{\dpix_{\indexOfVirtualCamera}}\geq0\comma
\end{equation}
 and $\ddb=-\frac{\translationVectorComp_{z}}{\dpix_{\indexOfVirtualCamera}}$
are independent from the depth of the considered point.

\subsubsection{Baseline: the probabilistic framework}

\label{subsec:Inverse-parallax-uncert}

As the aleatoric uncertainty is assumed to be proportional to the
variance of the estimated output distribution, the natural solution
to get the uncertainty on depth is to find the relation between the
variance of the parallax output distribution and the one of depth.
From the literature, we know that training a network as the log-likelihood
of the $L_{1}$ distance on the depth $\dpix$ makes the assumption
that its outputs follow a Laplace distribution whose mean and standard
deviation are respectively equal to $\hat{\location}\left(\dpix\right)$
and $\hat{\sigma}\left(\dpix\right)$. Unfortunately, the inverse
relation linking depth and parallax (see \Eqx{\ref{equ:disp_to_depth-simplified}})
means that a direct conversion between the variances, and therefore
the standard deviations, is impossible as they may not be finite in
both domains at the same time. However, training the network to infer
the inverse parallax solves this issue since injecting the variable
change $\zeta=\nicefrac{1}{\disp}$ in \Eqx{\ref{equ:disp_to_depth-simplified}}
gives:
\begin{equation}
\dpix=\dda\zeta+\ddb\Rightarrow\mathbb{\sigma}\left(\dpix\right)=\mathbb{\sigma}\left(\dda\zeta+\ddb\right)=\left|\dda\right|\mathbb{\sigma}\left(\zeta\right)=\dda\mathbb{\sigma}\left(\zeta\right)\pointMath
\end{equation}

In practice, we can train M4Depth to infer the uncertainty $\hat{\sigma}\left(\dpix\right)$
jointly to depth, from the inverse parallax by adding this term to
its loss function:
\begin{equation}
\mathcal{L}_{\dpix,t}=\frac{1}{\imHeight\imWidth}\sum_{\levelIndex=1}^{M}\sum_{\dpix_{ij}\in\dmap_{t}^{l}}2^{-\levelIndex}\ \left[\frac{\oslash\left(\left|\dpix_{ij}-\hat{\dpix}_{ij}\right|\right)}{\dda\hat{\sigma}\left(\zeta_{ij}\right)}+\beta\,\log\left(\dda\hat{\sigma}\left(\zeta_{ij}\right)\right)\right]\comma\label{eq:iwssip-loss-depth}
\end{equation}
where gradients are not propagated to the variables enclosed in the
$\oslash\left(\right)$ expression to avoid interference with the
gradients generated by the $\mathcal{L}_{t}$ term of the loss, and
where $\beta$ is an arbitrary weighting factor for the uncertainty
(we use $\beta=0.02$ in our experiments). Note that this loss is
only computed for pixels whose depth is lower than $400\,m$ to avoid
any convergence issues.

In the following, we will refer to our modified version of M4Depth
trained with this loss term as \mdudepth.

\begin{figure}
\begin{centering}
\resizebox{\columnwidth}{!}{%
\begin{tabular}{cc}
\includegraphics[viewport=0bp 10bp 241bp 231bp,clip,width=0.49\columnwidth]{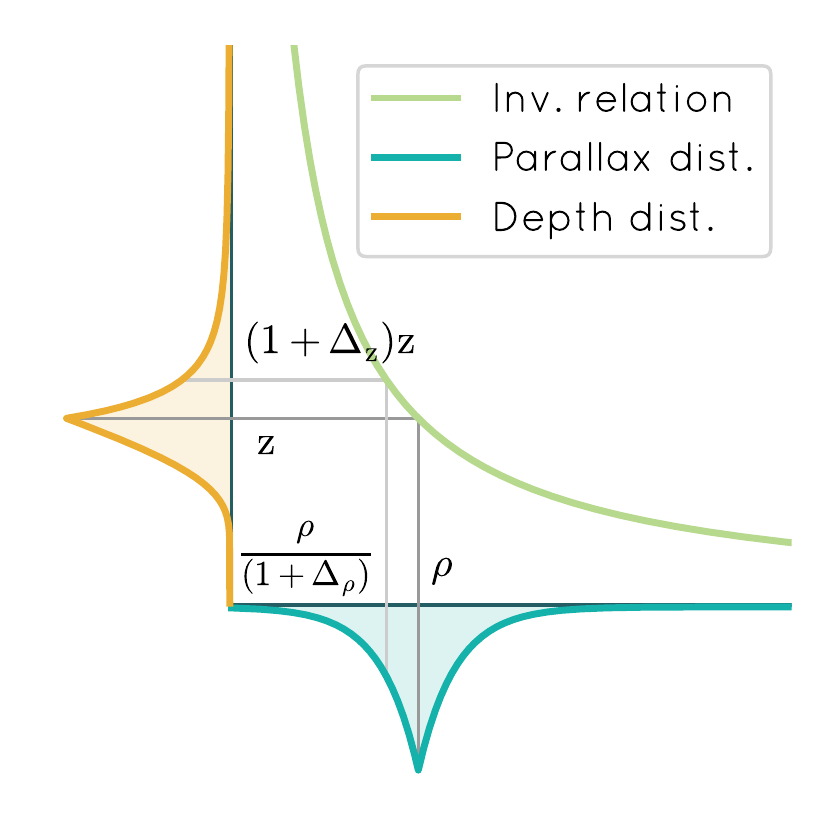} & \includegraphics[viewport=0bp 10bp 241bp 231bp,clip,width=0.49\columnwidth]{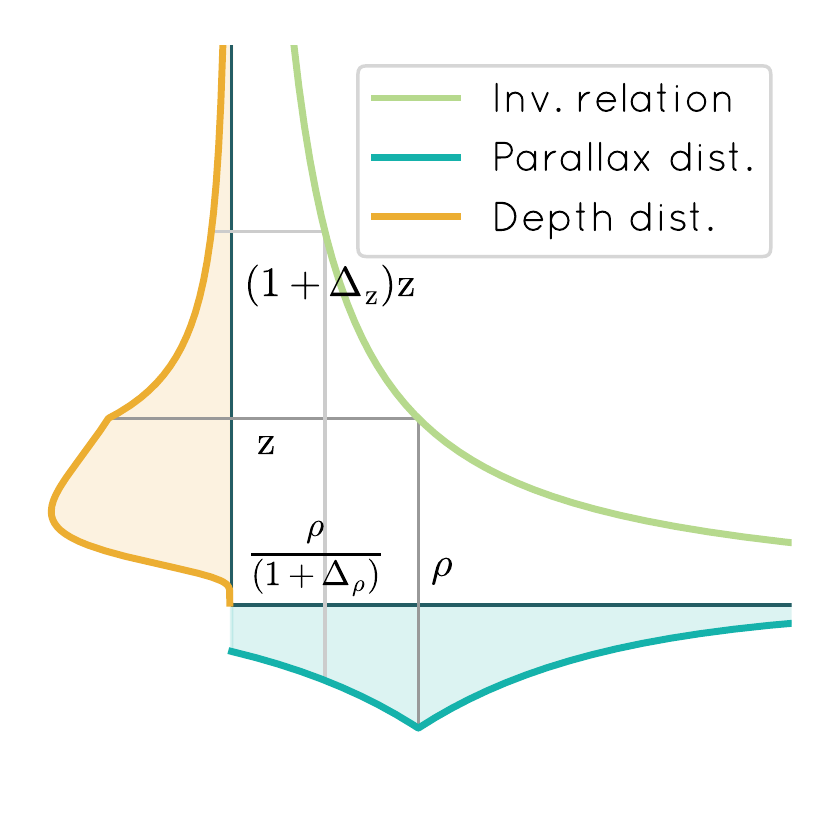}\tabularnewline
(a) Low standard deviation & (b) Large standard deviation\tabularnewline
\end{tabular}}
\par\end{centering}
\caption{Illustration of the correspondence between a Laplace distribution
(blue curve) and its inverse (orange curve) when applied to the relation
linking parallax to depth for different standard deviations of the
Laplace distribution. We propose to use $\Delta_{\protect\disp}$
as an uncertainty measure on the parallax, whose correspondence for
depth is $\Delta_{\protect\dpix}$. \label{fig:laplace_dist_illustration}}
\end{figure}

\subsubsection{Elaborate conversion of the uncertainty\label{subsec:Direct-parallax-uncert}}

One of the strengths of M4Depth is the direct link that exists between
the parallax and the disparity sweeping cost volumes, which are the
main sources of information available to infer the parallax. We assume
that, as the cost volumes provide valuable information on the parallax,
they should also provide valuable information on the related uncertainty.
This information is best used if there is a trivial relation between
the distribution to learn and the cost volumes. However, such a trivial
relation does not exist when learning the distribution of the inverse
parallax because of the inverse relation. As a result, the probabilistic
approach is not well suited for M4Depth, and we propose another approach
to get depth uncertainty estimates from the parallax domain.

Substituting the loss term $\mathcal{L}_{\dpix,t}$ defined in \Eqx{\ref{eq:iwssip-loss-depth}}
by the following $\mathcal{L}_{\disp,t}$ term in the training loss
of M4Depth allows to train the network to produce uncertainty estimates
$\hat{\sigma}\left(\disp\right)$ directly related to parallax estimates
$\hat{\disp}$:
\begin{equation}
\mathcal{L}_{\disp,t}=\frac{1}{\imHeight\imWidth}\sum_{\levelIndex=1}^{M}\sum_{\disp_{ij}\in\dispmap_{t}^{l}}2^{-\levelIndex}\ \left[\frac{\oslash\left(\left|\disp_{ij}-\hat{\disp}_{ij}\right|\right)}{\hat{\sigma}\left(\disp_{ij}\right)}+\beta\,\log\left(\hat{\sigma}\left(\disp_{ij}\right)\right)\right]\pointMath
\end{equation}
In the following, we will refer to our modified version of M4Depth
trained with this loss term as \mdupara.

We want to find a value $\Delta_{\dpix}>0$ in the depth domain that
represents the uncertainty carried by $\hat{\sigma}\left(\disp\right)$.
Stated otherwise, for any corresponding pair $\left(\sigma\left(\disp\right),\Delta_{\dpix}\right)$
and with everything else being equal, we want:
\begin{equation}
\sigma_{1}\left(\disp\right)<\sigma_{2}\left(\disp\right)\,\,\Leftrightarrow\,\,\Delta_{\dpix1}<\Delta_{\dpix2}\pointMath\label{eq:cond_uncert}
\end{equation}

Assuming that $\hat{\sigma}\left(\disp\right)$ is a valid indicator
for the uncertainty, we derive a notion of relative uncertainty $\Delta_{\disp}$
on the parallax defined as follows:
\begin{equation}
\Delta_{\disp}=\frac{\hat{\sigma}\left(\disp\right)}{\hat{\disp}}>0\pointMath
\end{equation}
This allows us to derive a range of values $\left[\frac{\hat{\disp}}{1+\Delta_{\disp}},\hat{\disp}\right]$
that is representative of the uncertainty as it monotonously increases
with uncertainty. As shown in \Figx{\ref{fig:laplace_dist_illustration}},
the equivalent of this range in the depth domain can be defined as
$\left[\hat{\dpix},\left(1+\Delta_{\dpix}\right)\hat{\dpix}\right]$
where $\hat{\dpix}=Z\left(\hat{\disp}\right)$. With this definition,
$\Delta_{\dpix}$ has properties similar to that of $\Delta_{\disp}$,
and it is also representative of the uncertainty on the parallax since
\Eqx{\ref{eq:cond_uncert}} is verified.

To find the relation between $\Delta_{\dpix}$ and $\Delta_{\disp}$,
we use \Eqx{\ref{equ:disp_to_depth-simplified}} as follows: 
\begin{equation}
\left(1+\Delta_{\dpix}\right)\hat{\dpix}=Z\left(\frac{\hat{\disp}}{1+\Delta_{\disp}}\right)\Leftrightarrow\Delta_{\dpix}=\frac{\ddb}{\hat{\dpix}}+\left(1+\Delta_{\disp}\right)\left(1-\frac{\ddb}{\hat{\dpix}}\right)-1>0\pointMath\label{eq:uncert_conv_good}
\end{equation}
Since $\Delta_{\disp}>0$ and $\hat{\dpix}>0$, the inequality is
verified if $\dpix_{\indexOfVirtualCamera}\hat{\dpix}+\translationVectorComp_{z}>0$
which is the same condition of existence than for the parallax itself~\cite{Fonder2022Parallax}.
Therefore, the $\Delta_{\dpix}$ quantity is defined for any possible
value of the parallax.

In a nutshell, getting joint depth and uncertainty estimates with
\mdupara amounts to training the network to infer the parallax and
its related uncertainty, then to convert them into depth $\dpix$
and its related uncertainty $\Delta_{\dpix}$ by using \Eqxy{\ref{equ:disp_to_depth-simplified}}{\ref{eq:uncert_conv_good}}
respectively.

\section{Experiments\label{sec:Experiments}}

In the experiments, we compare our elaborate approach for uncertainty
estimation to (1) the probabilistic baseline in various conditions,
and (2) existing methods on a benchmark for MVD methods. Before presenting
the results, let us first describe the experimental setup.

\subsection{Experimental setup}

\paragraph*{Datasets.}

We base our experiments on three datasets, namely \midair~\cite{Fonder2019MidAir},
\kitti~\cite{Geiger2012AreWe}, and \tartanair~\cite{Wang2020TartanAir}:
we use \midair to train and test the method in unstructured environments,
\kitti for zero-shot transfer tests on real data in urban environments,
and \tartanair for further tests either in urban or unstructured
environments. We use the same splits and image resolution as for the
original experiments for M4Depth~\cite{Fonder2022Parallax}.

\paragraph*{Performance evaluation.}

The performance analysis is based on a subset of metrics proposed
by Eigen~\etal~\cite{Eigen2014Depth} for depth estimation, that
is ``Abs rel'', ``RMSE log'', and $\delta<1.25$. We also report
the quality of uncertainty estimates with the \textit{Area under the
Sparsification Error} (AuSE) proposed Ilg~\etal~\cite{Ilg2018Uncertainty}.
This value, derived from so-called \textit{sparsification plots}~\cite{MacAodha2013Learning,Kondermann2007AnAdaptative,Wannenwetsch2017ProbFlow,Kybic2011Bootstrap},
has to be minimized for each performance metric for depth estimation.
Similar to related works, distant points (ground-truth depth~$>80\,m$)
are excluded from the performance metric computations.

\paragraph*{Network training.}

All the performance reported and analyzed in this section are based
on networks with six levels trained on the training set of the \midair
dataset. We use the same hyper-parameters and the same data augmentation
steps as the ones used for M4Depth~\cite{Fonder2022Parallax}. However,
we let the network train on more iterations ($250\,\text{k}$ steps).
We compute the performance of the network in validation after each
epoch and use the set of weights that performed the best in validation
for our performance analysis.

\subsection{Results}

\begin{table}
\begin{centering}
\caption{Performance of \mdupara compared to \mdudepth. \label{tab:Perf-comp-2-strats}}
\begin{adjustbox}{width=0.99\linewidth}%
\begin{tabular}{cccccccc}
\hline 
\multirow{2}{*}{Set} & \multirow{2}{*}{Method} & \multicolumn{2}{c}{Abs Rel} & \multicolumn{2}{c}{RMSE log} & \multicolumn{2}{c}{$\delta<1.25$}\tabularnewline
 &  & Perf. $\downarrow$ & AuSE $\downarrow$ & Perf. $\downarrow$ & AuSE $\downarrow$ & Perf. $\uparrow$ & AuSE $\downarrow$\tabularnewline
\hline 
\multirow{3}{*}{ \rotatebox[origin=c]{90}{{\midair}}} & M4Depth & 0.127 & $-$ & 0.185 & $-$ & 0.907 & $-$\tabularnewline
 & \mdudepth & 0.145 & 0.028 & 0.190 & 0.084 & 0.906 & 0.009\tabularnewline
 & \mdupara & 0.134 & \textbf{0.007} & 0.188 & \textbf{0.020} & 0.906 & \textbf{0.006}\tabularnewline
\hline 
\multirow{3}{*}{ \rotatebox[origin=c]{90}{{\kitti}}} & M4Depth & 0.193 & $-$ & 0.224 & $-$ & 0.849 & $-$\tabularnewline
 & \mdudepth & 0.140 & 0.025 & 0.195 & 0.046 & 0.858 & 0.021\tabularnewline
 & \mdupara & 0.147 & \textbf{0.021} & 0.195 & \textbf{0.041} & 0.858 & \textbf{0.019}\tabularnewline
\hline 
\multirow{3}{*}{ \rotatebox[origin=c]{90}{{TtA-W}}} & M4Depth & 0.614 & $-$ & 0.593 & $-$ & 0.652 & $-$\tabularnewline
 & \mdudepth & 0.618 & 0.176 & 0.597 & 0.217 & 0.636 & 0.031\tabularnewline
 & \mdupara & 0.478 & \textbf{0.058} & 0.592 & \textbf{0.157} & 0.646 & \textbf{0.028}\tabularnewline
\hline 
\multirow{3}{*}{ \rotatebox[origin=c]{90}{{TtA-O}}} & M4Depth & 0.446 & $-$ & 0.355 & $-$ & 0.793 & $-$\tabularnewline
 & \mdudepth & 0.468 & 0.077 & 0.410 & 0.155 & 0.776 & \textbf{0.020}\tabularnewline
 & \mdupara & 0.268 & \textbf{0.032} & 0.382 & \textbf{0.122} & 0.789 & \textbf{0.020}\tabularnewline
\hline 
\vspace{-0.2cm}
 &  &  &  &  &  &  & \tabularnewline
\end{tabular}\end{adjustbox}
\par\end{centering}
\noindent\begin{minipage}[t]{1\columnwidth}%
The network was trained and tested with the two loss functions on
the \midair dataset, and tested in zero shot transfer on the other
datasets. We used the seasons forest winter (TtA-W) and neighborhood
(TtA-N) environments of the \tartanair dataset. The best AuSEs for
each set are highlighted in bold.%
\end{minipage}
\end{table}

\begin{figure}
\noindent \begin{centering}
\includegraphics[viewport=8bp 8bp 482bp 230bp,clip,width=0.99\linewidth]{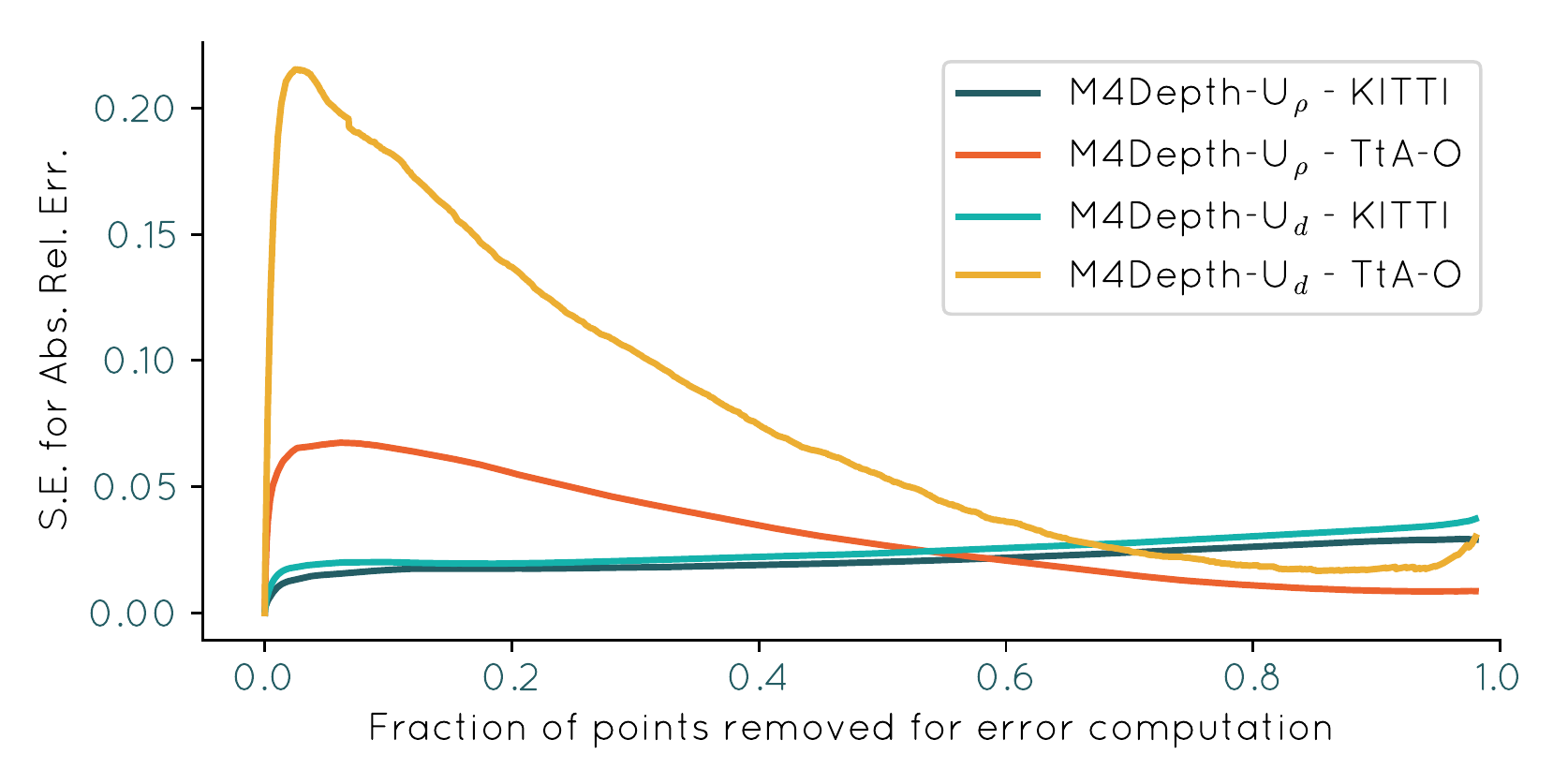}
\par\end{centering}
\caption{Sparsification error (S.E.) curves on the absolute relative error
for \mdudepth and \mdupara on the \kitti dataset, and on the ``Old
Town'' set of \tartanair (TtA-O).\label{fig:sparsification_error}}
\end{figure}

\paragraph*{\mdupara vs \mdudepth.}

In Section~\ref{sec:Uncertainty}, we explain how the probabilistic
framework can be used as a baseline, referred as \mdudepth, to get
the uncertainty on depth estimates for M4Depth. We also propose a
more elaborate method to get this uncertainty with \mdupara. We trained
our network with both methods on the \midair dataset, and tested
the performances in zero-shot transfer on various datasets. Some output
results for \mdupara are shown in \Figx{\ref{fig:graphical_abstract}}.

The results given in \tabx{\ref{tab:Perf-comp-2-strats}} and the
sparsification error curves displayed in \Figx{\ref{fig:sparsification_error}}
show that estimating depth and uncertainty jointly with M4Depth works
well. As hypothesized, the probabilistic framework underpinning the
\mdudepth baseline is sub-optimal while our elaborate uncertainty
conversion method, \mdupara, consistently performs better. Also,
the AuSE score for \mdupara varies less between datasets when compared
to \mdudepth, therefore hinting at more consistent generalization
performances. Finally, it is worth noting that both approaches for
estimating depth and its uncertainty preserve the raw performance
for depth estimation of M4Depth.

The sparsification error curves show that the uncertainty produced
by our network is better at discriminating low errors than higher
ones, since the sparsification error is higher for lower sparsification
values. The upward trend at the very end of the sparsification error
curve for \mdudepth on the \tartanair set hints that the network
is very confident in some areas with higher errors, which is not desired.
This behavior is not observed with \mdupara which further motivates
its interest over the baseline.

\begin{table}
\begin{centering}
\caption{Performance of \mdupara on the uncertainty benchmark proposed by
Schröppel~\etal~\cite{Schroppel2022ABenchmark} for MVS methods.
\label{tab:rmvd}}
\begin{adjustbox}{width=0.95\linewidth}%
\begin{tabular}{lcccc}
\hline 
Method & Causal & Abs. Rel. ($\downarrow$) & AuSE & Time {[}ms{]}\tabularnewline
\hline 
MVSNet~\cite{Yao2018mvsnet} & {\scriptsize{}\XSolidBrush{}} & 0.140 & 0.025 & 150\tabularnewline
Fast-MVSNet~\cite{Yu2020FastMVSNet} & {\scriptsize{}\XSolidBrush{}} & 0.121 & 0.034 & 350\tabularnewline
Vis-MVSNet~\cite{Zhang2022VisMVSNet} & {\scriptsize{}\XSolidBrush{}} & 0.103 & 0.028 & 820\tabularnewline
Robust MVD~\cite{Schroppel2022ABenchmark} & {\scriptsize{}\XSolidBrush{}} & 0.071 & 0.017 & 60\tabularnewline
\hline 
\mdupara & {\footnotesize{}\Checkmark{}} & 0.086 & 0.020 & 26\tabularnewline
\hline 
\vspace{-0.2cm}
 &  &  &  & \tabularnewline
\end{tabular}\end{adjustbox}
\par\end{centering}
\centering{}%
\noindent\begin{minipage}[t]{1\columnwidth}%
Performances are reported in zero-shot transfer on the 93-images test
set for the \kitti dataset used for this benchmark. Inference timings
are reported for full-size \kitti images. Note that \mdupara is
causal and only uses a sequence of frames that precedes the frame
considered for depth inference, while MVD methods are anti-causal
as they also use upcoming frames.%
\end{minipage}
\end{table}

\paragraph*{Robust MVD benchmark.}

As our method targets autonomous UAV applications, it has to produce
estimates for the latest available frame. Therefore, it cannot use
future information as opposed to generic multi-view depth estimation
methods which can use all the past and upcoming frames of the sequence.
Since we are the first to target this specific use case, there is
no existing baseline to compare to directly. Nonetheless, we assess
the value proposition of \mdupara over some other existing methods
on the benchmark proposed by Schröppel~\etal~\cite{Schroppel2022ABenchmark}
for joint multi-view depth and uncertainty estimation. Results on
the \kitti set of the benchmark are reported in \tabx{\ref{tab:rmvd}}.
Despite working with fewer data than other methods, \mdupara outperforms
most of the baseline and comes close to the state of the art on this
benchmark. This, combined with the fact that \mdupara is at least
$2.5$ times faster than other methods, leads us to conclude that
our method performs on par with existing methods tested on this benchmark,
and that \mdupara has a real benefit for practical use.

\paragraph*{Inference statistics.}

In the configuration used in our experiments, our method has $5.7\,\textrm{M}$
parameters, and requires up to $840\,\textrm{Mo}$ of VRAM to run.
On a NVidia V100 GPU and for input samples with a size of $384\times384$
pixels, \mdupara jointly estimates depth and uncertainty in $18\,\textrm{ms}$.
This is $1\,\textrm{ms}$ more than M4Depth, which means that estimating
the uncertainty requires a negligible additional computational time
when compared to depth estimation alone.

\section{Conclusion\label{sec:Conclusion}}

In this paper, we showed that it is possible to adapt M4Depth, an
efficient depth estimation network designed for autonomous vehicles
applications, for joint depth and uncertainty estimation at minimal
cost. We also demonstrated that converting the uncertainty values
produced by the network into uncertainty values related to depth is
better done with an elaborate conversion method, referred as \mdupara,
than with the standard probabilistic approach. The performance on
the \midair dataset and our tests in zero-shot transfer on the \kitti
and \tartanair datasets show that our method emerges as an excellent
joint depth and uncertainty estimator. In addition, testing \mdupara
on the Robust MVD benchmark in zero-shot transfer confirm that our
method performs similarly to other multi-view stereo methods, while
being $2.5$ times faster and causal, as opposed to these methods.

\section*{Acknowledgement}

This work was partly supported by the Walloon Region (Service Public
de Wallonie Recherche, Belgium) under grant n°2010235 (ARIAC by DigitalWallonia.ai).



\begin{thebibliography}{10}
\providecommand{\url}[1]{#1}
\csname url@samestyle\endcsname
\providecommand{\newblock}{\relax}
\providecommand{\bibinfo}[2]{#2}
\providecommand{\BIBentrySTDinterwordspacing}{\spaceskip=0pt\relax}
\providecommand{\BIBentryALTinterwordstretchfactor}{4}
\providecommand{\BIBentryALTinterwordspacing}{\spaceskip=\fontdimen2\font plus
\BIBentryALTinterwordstretchfactor\fontdimen3\font minus
  \fontdimen4\font\relax}
\providecommand{\BIBforeignlanguage}[2]{{%
\expandafter\ifx\csname l@#1\endcsname\relax
\typeout{** WARNING: IEEEtran.bst: No hyphenation pattern has been}%
\typeout{** loaded for the language `#1'. Using the pattern for}%
\typeout{** the default language instead.}%
\else
\language=\csname l@#1\endcsname
\fi
#2}}
\providecommand{\BIBdecl}{\relax}
\BIBdecl

\bibitem{Mumuni2022Deep}
\BIBentryALTinterwordspacing
F.~Mumuni, A.~Mumuni, and C.~K. Amuzuvi, ``Deep learning of monocular depth,
  optical flow and ego-motion with geometric guidance for {UAV} navigation in
  dynamic environments,'' \emph{Mach. Learn. Appl.}, vol.~10, pp. 2--15, Dec.
  2022. [Online]. Available: \url{https://doi.org/10.1016/j.mlwa.2022.100416}
\BIBentrySTDinterwordspacing

\bibitem{Yang2019Reactive}
\BIBentryALTinterwordspacing
X.~Yang, H.~Luo, Y.~Wu, Y.~Gao, C.~Liao, and K.-T. Cheng, ``Reactive obstacle
  avoidance of monocular quadrotors with online adapted depth prediction
  network,'' \emph{Neurocomputing}, vol. 325, pp. 142--158, Jan. 2019.
  [Online]. Available: \url{https://doi.org/10.1016/j.neucom.2018.10.019}
\BIBentrySTDinterwordspacing

\bibitem{Wang2020UAVEnvironmental}
\BIBentryALTinterwordspacing
D.~Wang, W.~Li, X.~Liu, N.~Li, and C.~Zhang, ``{UAV} environmental perception
  and autonomous obstacle avoidance: A deep learning and depth camera combined
  solution,'' \emph{Comput. Electron. Agric.}, vol. 175, pp. 1--11, Aug. 2020.
  [Online]. Available: \url{https://doi.org/10.1016/j.compag.2020.105523}
\BIBentrySTDinterwordspacing

\bibitem{Fonder2022Parallax}
\BIBentryALTinterwordspacing
M.~Fonder, D.~Ernst, and M.~Van~Droogenbroeck, ``Parallax inference for robust
  temporal monocular depth estimation in unstructured environments,''
  \emph{Sensors}, vol.~22, no.~23, pp. 1--22, Dec. 2022. [Online]. Available:
  \url{https://doi.org/10.3390/s22239374}
\BIBentrySTDinterwordspacing

\bibitem{Fonder2019MidAir}
\BIBentryALTinterwordspacing
M.~Fonder and M.~Van~Droogenbroeck, ``Mid-air: A multi-modal dataset for
  extremely low altitude drone flights,'' in \emph{IEEE Int. Conf. Comput. Vis.
  Pattern Recognit. Work. (CVPRW), UAVision}.\hskip 1em plus 0.5em minus
  0.4em\relax Long Beach, CA, USA: Inst. Electr. Electron. Eng. (IEEE), Jun.
  2019, pp. 553--562. [Online]. Available:
  \url{https://doi.org/10.1109/CVPRW.2019.00081}
\BIBentrySTDinterwordspacing

\bibitem{Geiger2012AreWe}
\BIBentryALTinterwordspacing
A.~Geiger, P.~Lenz, and R.~Urtasun, ``Are we ready for autonomous driving?
  {T}he {KITTI} vision benchmark suite,'' in \emph{IEEE Int. Conf. Comput. Vis.
  Pattern Recognit. (CVPR)}, Providence, RI, USA, Jun. 2012, pp. 3354--3361.
  [Online]. Available: \url{https://doi.org/10.1109/CVPR.2012.6248074}
\BIBentrySTDinterwordspacing

\bibitem{Gawlikowski2021ASurvey}
\BIBentryALTinterwordspacing
J.~Gawlikowski, C.~R.~N. Tassi, M.~Ali, J.~Lee, M.~Humt, J.~Feng, A.~Kruspe,
  R.~Triebel, P.~Jung, R.~Roscher, M.~Shahzad, W.~Yang, R.~Bamler, and X.~X.
  Zhu, ``A survey of uncertainty in deep neural networks,'' \emph{CoRR}, vol.
  abs/2107.03342, 2021. [Online]. Available:
  \url{https://doi.org/10.48550/arXiv.2107.03342}
\BIBentrySTDinterwordspacing

\bibitem{Kendall2017What}
A.~Kendall and Y.~Gal, ``What uncertainties do we need in {Bayesian} deep
  learning for computer vision?'' in \emph{Adv. Neural Inf. Process. Syst.
  (NeurIPS)}, Long Beach, CA, USA, Dec. 2017, pp. 5574--5584.

\bibitem{Ke2021Deep}
\BIBentryALTinterwordspacing
T.~Ke, T.~Do, K.~Vuong, K.~Sartipi, and S.~I. Roumeliotis, ``Deep multi-view
  depth estimation with predicted uncertainty,'' in \emph{IEEE Int. Conf.
  Robot. Autom. (ICRA)}.\hskip 1em plus 0.5em minus 0.4em\relax Xian, China:
  Inst. Electr. Electron. Eng. (IEEE), May 2021, pp. 9235--9241. [Online].
  Available: \url{https://doi.org/10.1109/ICRA48506.2021.9560873}
\BIBentrySTDinterwordspacing

\bibitem{Ilg2018Uncertainty}
\BIBentryALTinterwordspacing
E.~Ilg, {\"O}.~{\c C}i{\c c}ek, S.~Galesso, A.~Klein, O.~Makansi, F.~Hutter,
  and T.~Brox, ``Uncertainty estimates and multi-hypotheses networks for
  optical flow,'' in \emph{Eur. Conf. Comput. Vis. (ECCV)}, ser. Lect. Notes
  Comput. Sci., vol. 11211.\hskip 1em plus 0.5em minus 0.4em\relax Springer
  Int. Publ., 2018, pp. 677--693. [Online]. Available:
  \url{https://doi.org/10.1007/978-3-030-01234-2_40}
\BIBentrySTDinterwordspacing

\bibitem{Schroppel2022ABenchmark}
\BIBentryALTinterwordspacing
P.~Schr{\"o}ppel, J.~Bechtold, A.~Amiranashvili, and T.~Brox, ``A benchmark and
  a baseline for robust multi-view depth estimation,'' \emph{CoRR}, vol.
  abs/2209.06681, 2022. [Online]. Available:
  \url{https://doi.org/10.48550/arXiv.2209.06681}
\BIBentrySTDinterwordspacing

\bibitem{Zhang2022VisMVSNet}
\BIBentryALTinterwordspacing
J.~Zhang, S.~Li, Z.~Luo, T.~Fang, and Y.~Yao, ``Vis-{MVSNet}: Visibility-aware
  multi-view stereo network,'' \emph{Int. J. Comput. Vis.}, vol. 131, no.~1,
  pp. 199--214, Oct. 2022. [Online]. Available:
  \url{https://doi.org/10.1007/s11263-022-01697-3}
\BIBentrySTDinterwordspacing

\bibitem{Poggi2020On}
\BIBentryALTinterwordspacing
M.~Poggi, F.~Aleotti, F.~Tosi, and S.~Mattoccia, ``On the uncertainty of
  self-supervised monocular depth estimation,'' in \emph{IEEE/CVF Conf. Comput.
  Vis. Pattern Recognit. (CVPR)}.\hskip 1em plus 0.5em minus 0.4em\relax
  Seattle, WA, USA: Inst. Electr. Electron. Eng. (IEEE), Jun. 2020, pp.
  3224--3234. [Online]. Available:
  \url{https://doi.org/10.1109/cvpr42600.2020.00329}
\BIBentrySTDinterwordspacing

\bibitem{Su2022Uncertainty}
\BIBentryALTinterwordspacing
W.~Su, Q.~Xu, and W.~Tao, ``Uncertainty guided multi-view stereo network for
  depth estimation,'' \emph{IEEE Trans. Circuits Syst. Video Technol.},
  vol.~32, no.~11, pp. 7796--7808, Nov. 2022. [Online]. Available:
  \url{https://doi.org/10.1109/TCSVT.2022.3183836}
\BIBentrySTDinterwordspacing

\bibitem{Liu2019Neural}
\BIBentryALTinterwordspacing
C.~Liu, J.~Gu, K.~Kim, S.~G. Narasimhan, and J.~Kautz, ``Neural {RGB-D}
  sensing: Depth and uncertainty from a video camera,'' in \emph{IEEE/CVF Conf.
  Comput. Vis. Pattern Recognit. (CVPR)}.\hskip 1em plus 0.5em minus
  0.4em\relax Long Beach, CA, USA: Inst. Electr. Electron. Eng. (IEEE), Jun.
  2019, pp. 10\,978--10\,987. [Online]. Available:
  \url{https://doi.org/10.1109/CVPR.2019.01124}
\BIBentrySTDinterwordspacing

\bibitem{Zhao2021AConfidenceBased}
\BIBentryALTinterwordspacing
W.~Zhao, S.~Liu, Y.~Wei, H.~Guo, and Y.-J. Liu, ``A confidence-based iterative
  solver of depths and surface normals for deep multi-view stereo,'' in
  \emph{IEEE Int. Conf. Comput. Vis. (ICCV)}.\hskip 1em plus 0.5em minus
  0.4em\relax Montreal, QC, Canada: Inst. Electr. Electron. Eng. (IEEE), Oct.
  2021, pp. 6148--6157. [Online]. Available:
  \url{https://doi.org/10.1109/iccv48922.2021.00611}
\BIBentrySTDinterwordspacing

\bibitem{Mehltretter2021Aleatoric}
\BIBentryALTinterwordspacing
M.~Mehltretter and C.~Heipke, ``Aleatoric uncertainty estimation for dense
  stereo matching via {CNN}-based cost volume analysis,'' \emph{ISPRS J.
  Photogramm. Remote Sens.}, vol. 171, pp. 63--75, Jan. 2021. [Online].
  Available: \url{https://doi.org/10.1016/j.isprsjprs.2020.11.003}
\BIBentrySTDinterwordspacing

\bibitem{Homeyer2022Multiview}
\BIBentryALTinterwordspacing
C.~Homeyer, O.~Lange, and C.~Schn{\"o}rr, ``Multi-view monocular depth and
  uncertainty prediction with deep {SfM} in dynamic environments,'' in
  \emph{Int. J. Pattern Recognit. Artif. Intell.}, ser. Lect. Notes Comput.
  Sci., vol. 13363.\hskip 1em plus 0.5em minus 0.4em\relax Springer Int. Publ.,
  2022, pp. 373--385. [Online]. Available:
  \url{https://doi.org/10.1007/978-3-031-09037-0_31}
\BIBentrySTDinterwordspacing

\bibitem{Klodt2018Supervising}
\BIBentryALTinterwordspacing
M.~Klodt and A.~Vedaldi, ``Supervising the new with the old: Learning {SFM}
  from {SFM},'' in \emph{Eur. Conf. Comput. Vis. (ECCV)}, ser. Lect. Notes
  Comput. Sci., vol. 11214.\hskip 1em plus 0.5em minus 0.4em\relax Springer
  Int. Publ., 2018, pp. 713--728. [Online]. Available:
  \url{https://doi.org/10.1007/978-3-030-01249-6_43}
\BIBentrySTDinterwordspacing

\bibitem{Yang2021Fast}
\BIBentryALTinterwordspacing
X.~Yang, J.~Chen, Y.~Dang, H.~Luo, Y.~Tang, C.~Liao, P.~Chen, and K.-T. Cheng,
  ``Fast depth prediction and obstacle avoidance on a monocular drone using
  probabilistic convolutional neural network,'' \emph{IEEE Trans. Intell.
  Transp. Syst.}, vol.~22, no.~1, pp. 156--167, Jan. 2021. [Online]. Available:
  \url{https://doi.org/10.1109/TITS.2019.2955598}
\BIBentrySTDinterwordspacing

\bibitem{Wang2020TartanAir}
\BIBentryALTinterwordspacing
W.~Wang, D.~Zhu, X.~Wang, Y.~Hu, Y.~Qiu, C.~Wang, Y.~Hu, A.~Kapoor, and
  S.~Scherer, ``{TartanAir}: A dataset to push the limits of visual {SLAM},''
  in \emph{IEEE/RSJ Int. Conf. Intell. Robot. Syst. (IROS)}.\hskip 1em plus
  0.5em minus 0.4em\relax Las Vegas, NV, USA: Inst. Electr. Electron. Eng.
  (IEEE), Oct. 2020, pp. 4909--4916. [Online]. Available:
  \url{https://doi.org/10.1109/IROS45743.2020.9341801}
\BIBentrySTDinterwordspacing

\bibitem{Eigen2014Depth}
D.~Eigen, C.~Puhrsch, and R.~Fergus, ``Depth map prediction from a single image
  using a multi-scale deep network,'' in \emph{Adv. Neural Inf. Process. Syst.
  (NeurIPS)}, 2014, pp. 2366--2374.

\bibitem{MacAodha2013Learning}
\BIBentryALTinterwordspacing
O.~Mac~Aodha, A.~Humayun, M.~Pollefeys, and G.~J. Brostow, ``Learning a
  confidence measure for optical flow,'' \emph{IEEE Trans. Pattern Anal. Mach.
  Intell.}, vol.~35, no.~5, pp. 1107--1120, May 2013. [Online]. Available:
  \url{https://doi.org/10.1109/TPAMI.2012.171}
\BIBentrySTDinterwordspacing

\bibitem{Kondermann2007AnAdaptative}
\BIBentryALTinterwordspacing
C.~Kondermann, D.~Kondermann, B.~J{\"a}hne, and C.~Garbe, ``An adaptive
  confidence measure for optical flows based on linear subspace projections,''
  in \emph{Pattern Recognit.}, ser. Lect. Notes Comput. Sci., vol. 4713.\hskip
  1em plus 0.5em minus 0.4em\relax Springer, 2007, pp. 132--141. [Online].
  Available: \url{https://doi.org/10.1007/978-3-540-74936-3_14}
\BIBentrySTDinterwordspacing

\bibitem{Wannenwetsch2017ProbFlow}
\BIBentryALTinterwordspacing
A.~S. Wannenwetsch, M.~Keuper, and S.~Roth, ``{ProbFlow}: Joint optical flow
  and uncertainty estimation,'' in \emph{IEEE Int. Conf. Comput. Vis.
  (ICCV)}.\hskip 1em plus 0.5em minus 0.4em\relax Venice, Italy: Inst. Electr.
  Electron. Eng. (IEEE), Oct. 2017, pp. 1182--1191. [Online]. Available:
  \url{https://doi.org/10.1109/ICCV.2017.133}
\BIBentrySTDinterwordspacing

\bibitem{Kybic2011Bootstrap}
\BIBentryALTinterwordspacing
J.~Kybic and C.~Nieuwenhuis, ``Bootstrap optical flow confidence and
  uncertainty measure,'' \emph{Comput. Vis. Image Underst.}, vol. 115, no.~10,
  pp. 1449--1462, Oct. 2011. [Online]. Available:
  \url{https://doi.org/10.1016/j.cviu.2011.06.008}
\BIBentrySTDinterwordspacing

\bibitem{Yao2018mvsnet}
\BIBentryALTinterwordspacing
Y.~Yao, Z.~Luo, S.~Li, T.~Fang, and L.~Quan, ``{MVSNet}: Depth inference for
  unstructured multi-view stereo,'' in \emph{Eur. Conf. Comput. Vis. (ECCV)},
  ser. Lect. Notes Comput. Sci., vol. 11212.\hskip 1em plus 0.5em minus
  0.4em\relax Springer, 2018, pp. 785--801. [Online]. Available:
  \url{https://doi.org/10.1007/978-3-030-01237-3_47}
\BIBentrySTDinterwordspacing

\bibitem{Yu2020FastMVSNet}
\BIBentryALTinterwordspacing
Z.~Yu and S.~Gao, ``Fast-{MVSNet}: Sparse-to-dense multi-view stereo with
  learned propagation and gauss-newton refinement,'' in \emph{IEEE/CVF Conf.
  Comput. Vis. Pattern Recognit. (CVPR)}.\hskip 1em plus 0.5em minus
  0.4em\relax Seattle, WA, USA: Inst. Electr. Electron. Eng. (IEEE), Jun. 2020,
  pp. 1946--1955. [Online]. Available:
  \url{https://doi.org/10.1109/cvpr42600.2020.00202}
\BIBentrySTDinterwordspacing

\end{thebibliography}
\end{document}